\RequirePackage{fix-cm}
\documentclass[twocolumn]{svjour3}          
\smartqed  
\usepackage{graphicx}
\usepackage{subfigure}
\usepackage{cite}
\usepackage{mathptmx} 
\usepackage{balance} 
\usepackage{lipsum}
\usepackage[hyphens]{url}
\usepackage[hidelinks]{hyperref}
\usepackage{flushend}
\usepackage{latexsym}
\usepackage{amsmath}
\usepackage{blindtext}

\usepackage{natbib}
\setcitestyle{authoryear,round}

\usepackage{tikz,xcolor,hyperref}

\definecolor{lime}{HTML}{A6CE39}

\DeclareRobustCommand{\orcidicon}{
	\begin{tikzpicture}
	\draw[lime, fill=lime] (0,0) 
	circle [radius=0.16] 
	node[white] {{\fontfamily{qag}\selectfont \tiny ID}};
	\draw[white, fill=white] (-0.0625,0.095) 
	circle [radius=0.007];
	\end{tikzpicture}
	\hspace{-2mm}
}

\foreach \x in {A, ..., Z}{%
\expandafter\xdef\csname orcid\x\endcsname{\noexpand\href{https://orcid.org/\csname orcidauthor\x\endcsname}{\noexpand\orcidicon}}
}

\usepackage[misc]{ifsym}
\newcommand\blfootnote[1]{%
    \begingroup
    \renewcommand\thefootnote{}\footnote{#1}%
    \addtocounter{footnote}{-1}%
    \endgroup
}
\usepackage[hang]{footmisc}

\begin{document}

\title{Complete Solution for Vehicle Re-ID in Surround-view Camera System
}


\author{Zizhang Wu$^{1}$ \and Tianhao Xu$^{2}$\hspace{-1.5mm}\orcidA{} \and Fan Wang$^{1}$ \and Xiaoquan Wang$^{1}$ \and Jing Song$^{1}$}



\date{Received: date / Accepted: date}

\twocolumn[
\maketitle
\begin{@twocolumnfalse}

\begin{abstract}
Vehicle re-identification (Re-ID) is a critical component of the autonomous driving perception system, and research in this area has accelerated in recent years. However, there is yet no perfect solution to the vehicle re-identification issue associated with the car's surround-view camera system. Our analysis identifies two significant issues in the aforementioned scenario: i) It is difficult to identify the same vehicle in many picture frames due to the unique construction of the fisheye camera. ii) The appearance of the same vehicle when seen via the surround vision system's several cameras is rather different. To overcome these issues, we suggest an integrative vehicle Re-ID solution method. On the one hand, we provide a technique for determining the consistency of the tracking box drift with respect to the target. On the other hand, we combine a Re-ID network based on the attention mechanism with spatial limitations to increase performance in situations involving multiple cameras. Finally, our approach combines state-of-the-art accuracy with real-time performance. We will soon make the source code and annotated fisheye dataset available.
\end{abstract}

\end{@twocolumnfalse}
]

{
    \blfootnote{
        \hspace{-0.25in}\Letter  \quad Zizhang Wu
        \newline  zizhang.wu@zongmutech.com
        \newline
        }
    \blfootnote{
        \hspace{-0.25in}\Letter  \quad Tianhao Xu
        \newline  tianhao.xu@tu-braunschweig.de
        \newline
        }
        
    \footnotetext[1]{Zongmu Technology, Shanghai, China}
    \footnotetext[2]{Technical University of Braunschweig, Braunschweig, Germany}
}

\maketitle


\section{Introduction}

With the advent of autonomous driving, significant efforts have been devoted in the computer vision community to vehicle-related research. Especially, vehicle Re-ID is one of the most active research fields aiming at identifying the same vehicle across the image archive captured from different cameras. However, vehicle Re-ID in real-world scenarios is still a challenging task in the computer vision field despite its long success. Consequently, it is desirable to seek an effective and robust vehicle Re-ID method, which is the prerequisite for achieving the trajectory prediction, state estimation, and speed estimation of the target vehicle.

\hspace*{\fill}
~\\

Existing Re-ID studies mainly focus on the pedestrian Re-ID and the vehicle Re-ID. Unlike pedestrian Re-ID \cite{article1}, \cite{article2}, \cite{article3} which extract rich features from images with different poses and colors, vehicle Re-ID faces more severe challenges. Vehicles captured by cameras suffer from distortion, occlusion,  individual similarity, etc. Although the pioneering deep learning-based methods \cite{article14}, \cite{article28}, \cite{article29}, \cite{article30} can learn global semantic features from an entire vehicle image, they still fail to distinguish some small but discriminative regions.
Therefore, \cite{article4}, \cite{article5}, \cite{article6} utilized Spatio-temporal relations to learn the similarity between vehicle images, which boost the performance of Re-ID. However, the Spatio-temporal information is not annotated in all existing datasets, so some restrictions exist for exploring these methods. To improve the adaptability of vehicle Re-ID methods to different scenarios, \cite{article7}, \cite{article8}, \cite{article31} employ vehicle attributes to learn the correlations among vehicles. Although the vehicle attributes are more general and discriminating than other features, the relationship between vehicle attributes and categories is ignored. Subsequently, \cite{article9} introduced the attention mechanism in the attribute branch. This operation is conducive to selecting attributes for corresponding input vehicle images, which helps the category branch select better discriminative features for category recognition.

\begin{figure}[!t]
    \centering
    \includegraphics[width=7.5cm]{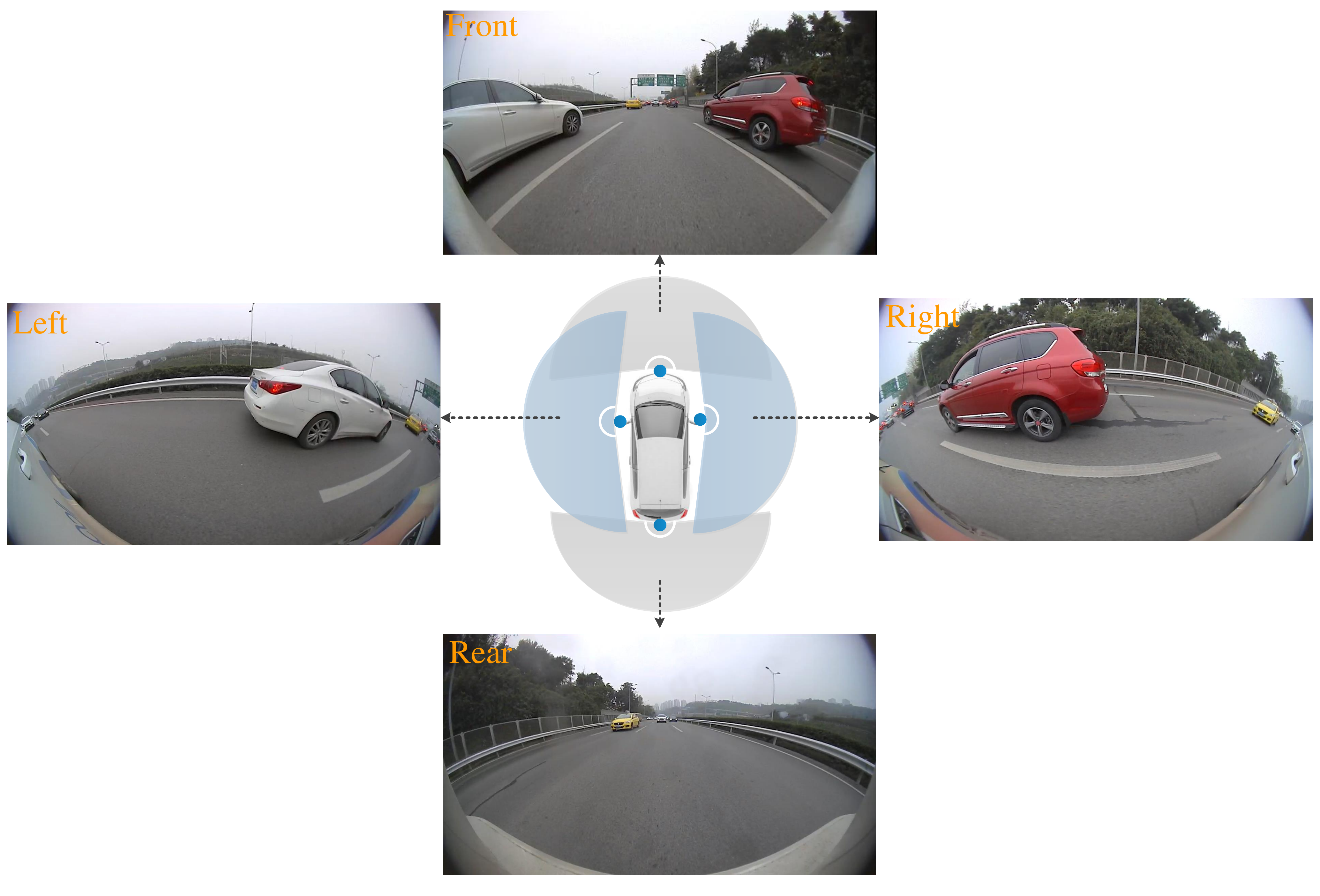}
    \caption{The inductance of oscillation winding on amorphous magnetic core versus DC bias magnetic field.}
    \label{fig:Figure01}
\end{figure}

Despite the tremendous progress in current vehicle Re-ID methods, most are designed for surveillance scenarios. While in autonomous driving, surround-view camera systems have become more and more popular. To achieve a 360$^{\circ}$ blind-spot free environment perception, we mount multiple fisheye cameras around the vehicle, as shown in Fig.\ref{fig:Figure01}. Therefore, it is essential to learn the relationship between the same vehicle among different vehicle-mounted cameras and different frames in a single camera. We state two main challenges for achieving a vehicle Re-ID solution in the surround-view multi-camera scenario: 

1) In a single-camera view, vehicle features in consecutive frames vary dramatically due to fisheye distortion, occlusion, truncation, etc. It is difficult to recognize the same vehicle from the past image archive under such interference.

2) In a multi-camera view, the appearance of the same vehicle varies significantly from different viewpoints. The individual similarity of vehicles also led to great confusion in matching.

In this paper, we propose our methods respectively to these challenges. For the balance of accuracy and efficiency,  SiamRPN++ network \cite{article10} and BDBnet \cite{article11} are employed for vehicle  re-identification in single-camera and multi-camera, respectively. However, these two models are mainly designed for surveillance scenes and are unable to deal with the significant variations in vehicle appearance in the fisheye system. Therefore, a post-process module named quality evaluation mechanism for the output bounding box is proposed to alleviate the target drift caused by fisheye distortion, occlusion, etc. Besides, an attention module and a spatial constraint strategy are introduced to respond to the intra-class difference and inter-class similarity of vehicles \cite{article7}. To drive the study of vehicle Re-ID in surround-view scenarios and fill the gap in the relative dataset, we will release the large-scale annotated fisheye dataset.

Our contributions can be summarized as follows:

\begin{itemize}

\item We provide an integrated vehicle Re-ID solution for the multi-camera surround-view scenario.
\item We propose a technique for evaluating the output bounding box's quality, which can alleviate the issue of target drift.
\item A novel spatial constraint strategy is introduced for regularizing the Re-ID results in the surround-view camera system.
\item A large-scale fisheye dataset with annotations is provided to aid in promoting relevant research.
\end{itemize}

\section{RELATED WORK AND OUR CONTRIBUTIONS}

This section summarises the literature on vehicle re-identifica\\tion techniques and tracking algorithms, both of which are closely connected to our study.

\paragraph{Vehicle Re-ID} Vehicle Re-ID has been widely studied in recent years. As well-known, the common challenge is how to deal with the inter-class similarity, and the intra-class difference \cite{article7}. Different vehicles have a similar appearance, while the same vehicle looks different due to the diverse perspectives and distortion. Until now, lots of work have been proposed to tackle this challenge.  \cite{article7} designed a pipeline, which adopts deep relative distance learning (DRDL) to project vehicle images into Euclidean space, then calculate the distance in Euclidean space to measure the similarity of two-vehicle images. \cite{article4} created a dataset named VeRi-776, which employs visual features, license plates, and spatial-temporal information to explore the Re-ID task of vehicles. Meanwhile, further works \cite{article5}, \cite{article6}, \cite{article12}, \cite{article33}, \cite{article32} introduced spatial-temporal information to boost the performance of Re-ID. \cite{article5} proposed a two-stage framework, which utilizes complex spatial-temporal information of vehicles to regularize Re-ID results effectively. Subsequently, \cite{article6} used spatial prior knowledge to generate the tracklet and selected the vehicle in the middle frame as a feature of the tracklet. Furthermore, \cite{article12} utilized location information between cameras to improve the accuracy of Re-ID. In light of the success of using location information, we exploit a novel spatial constraint strategy to enhance the Re-ID results in the surround-view camera system.

Since the spatial-temporal information in datasets is not always available, approaches for Re-ID have been proposed by combining local and global features of targets \cite{article13}, \cite{article11}, \cite{article14}, \cite{article15}, \cite{article34}, \cite{article35}. For instance, \cite{article13} designed a network with the combination of the local and global branches, which employs joint feature vectors to enhance robustness on occlusion problems. Different from the DropBlock, \cite{article11} proposed Batch DropBlock that drops the same region in a batch of the image to accomplish the metric learning task better. \cite{article14} enabled the model to perform intensive learning of local features from the loss function aspect. Inspired by the attention mechanism, methods of processing the local and global information are more flexible \cite{article36}, \cite{article37}, \cite{article38}, \cite{article39}, \cite{article40}. In this paper, we also utilize the attention mechanism to make the model focus on target regions and use triplet loss \cite{article15} and softmax loss to enhance the performance of vehicle Re-ID.

In some complex scenarios, the relative positions between vehicles are constantly changing, so the real-time performance of the model is also critical. \cite{chu2019vehicle} proposed a perspective-aware metric learning method for extreme viewpoint variations, in which a viewpoint-aware network (VANet) learned two metrics for comparable and dissimilar perspectived using two feature spaces, respectively. \cite{chen2020orientation} proposed a Semantics-guided Part Attention Network (SPAN) using semantic labels to predict attention masks at different angles of the vehicle to extract discriminative features for each component. \cite{liu2019urban} considered that pairs of vehicles differ visually in large intervals and proposed a Self-Attention Stair Feature Fusion model to learn discriminative features to capture the details of the images. In addition to the recognition difficulties caused by the viewing angle, lighting is also an important factor affecting accuracy. \cite{ma2019vehicle} proposed a refined part model to learn feature embeddings to automatically localized vehicles through a Grid Spatial Transformer Network (GSTN). The above methods and our proposed model are based on deep learning methods. Therefore, we performed a real-time performance comparison. \cite{chu2019vehicle} \cite{chen2020orientation} \cite{meng2020parsing} show the FPS of the model. In addition, the amount of parameter \cite{liu2019urban} \cite{ma2019vehicle} \cite{shen2021exploring} of the model is crucial in the application, because the arithmetic power of the electronic control unit(ECU) is limited.

\begin{figure}[thpb]
   \centering
   \parbox{3.3in}{		
        \begin{minipage}{1\linewidth}
			\centering
			\centerline{\includegraphics[width=1\linewidth]{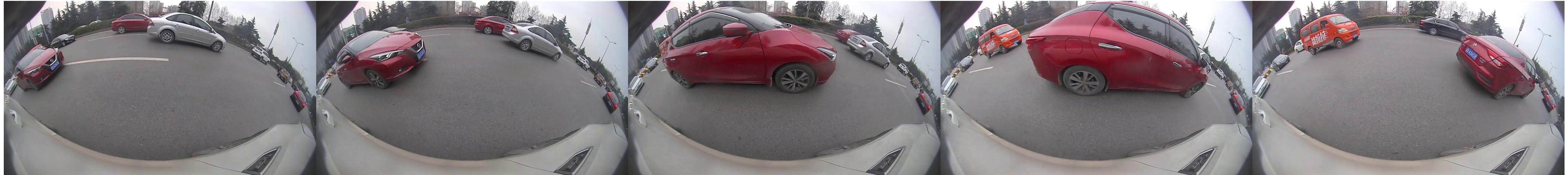}}
			\centerline{(a)}
		\end{minipage}
		
		\begin{minipage}{1\linewidth}
			\centering
			\centerline{\includegraphics[width=1\linewidth]{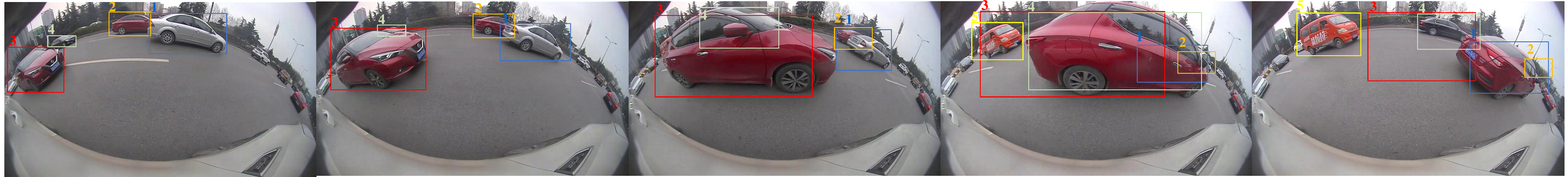}}
			\centerline{(b)}
		\end{minipage}
		
		\begin{minipage}{1\linewidth}
			\centering
			\centerline{\includegraphics[width=1\linewidth]{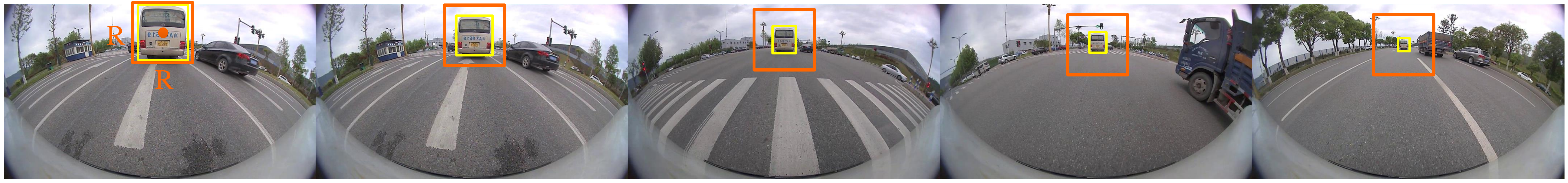}}
			\centerline{(c)}
		\end{minipage}
}
   
   \caption{Vehicles in the single view of a fisheye camera. (a) The same vehicle features change dramatically in consecutive frames and vehicles tend to obscure each other. (b) Matching errors are caused by tracking results. (c) The vehicle center indicated by the orange box is stable, while the IoU in consecutive frames indicated by the yellow box decreases with movement.}
   \label{fig:Figure02}
\end{figure}

\begin{figure*}[!h]
   \centering
   \parbox{6.9in}{\centering\includegraphics[width=1\linewidth]{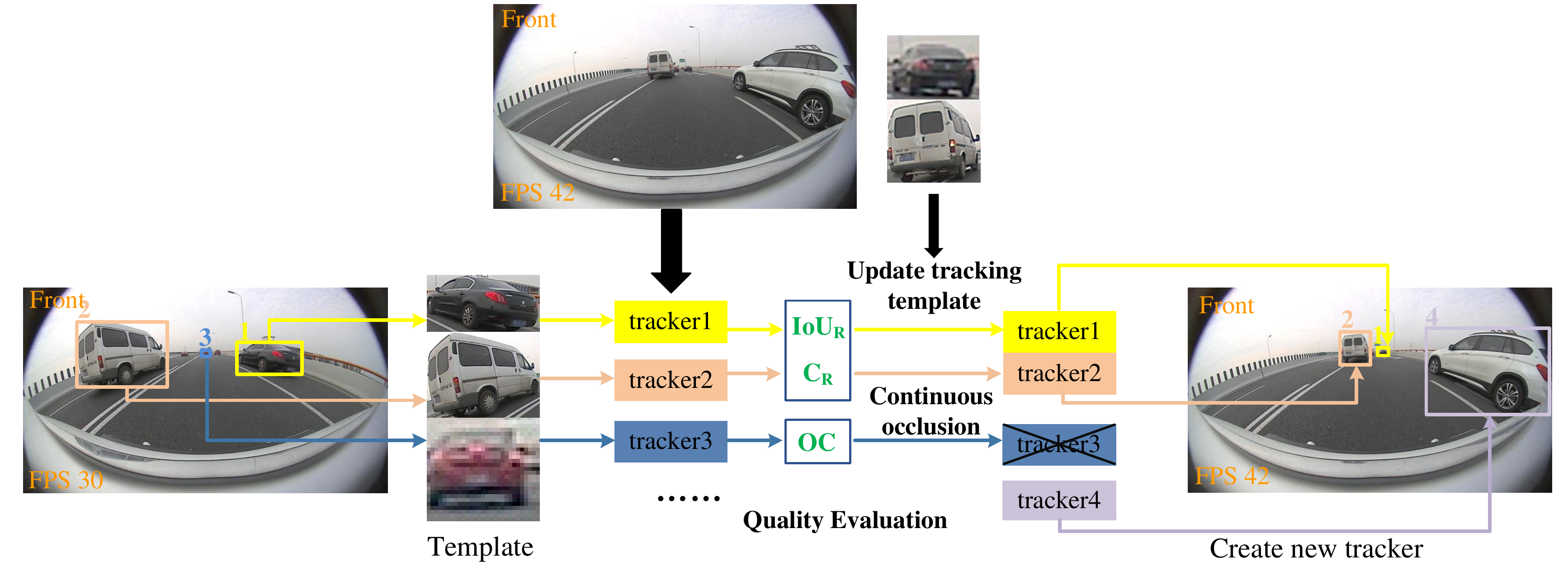}
}
   
   \caption{The overall framework of vehicle Re-ID in a single camera. Each object is assigned a single tracker to realize Re-ID in a single channel. Tracking templates are initialized with object detection results. All tracking outputs are post-processed by the quality evaluation module to deal with distorted or occluded objects.}
    \label{fig:Figure04}
\end{figure*}

\paragraph{Tracking algorithms related with Vehicle Re-ID}  Object tracking algorithm plays an important role in the implementation of vehicle Re-ID scheme \cite{article16}, \cite{article17}, \cite{article18}, \cite{article19}. For object tracking tasks, the trackers based on Siamese network \cite{article20},\cite{article21},\cite{article22},\cite{article23},\cite{article24} have received signiﬁcant attentions for their well-balanced tracking accuracy and efﬁciency, which is of great value for boosting vehicle Re-ID task in the surround-view multi-camera scenario. The seminal work \cite{article20}, \cite{article22}, \cite{article21} formulated visual tracking as a cross-correlation problem and produced a tracking similarity map from the depth model with the Siamese network, to find the location of the target by comparing the similarity between the target and the search region. However, these works have an inherent drawback, as their tracking accuracies on the OTB benchmark \cite{article25} still leave a relatively large gap with state-of-the-art deep trackers like ECO \cite{article26} and MDNet \cite{article27}. To overcome this drawback, \cite{article10} transferred the object position in the bounding box during the training phase to avoid the location bias of the network. Thus, the network can focus on the object marginal area of the search region, and the accuracy of Siamese trackers is boosted signiﬁcantly. Besides, this approach also proposed a lightweight depth-wise cross-correlation layer to improve the running speed. We adopt the SiamRPN++ model to realize the vehicle Re-ID task in a single camera considering its great performance.

\section{Vehicle Re-ID in surround-view camera system}
We separate the vehicle Re-ID task into two subtasks in the construction of the surround-view multi-camera system: single-camera Re-ID and multi-camera Re-ID. This section will describe our methodologies in detail for each subtask. The parameters in the method are selected based on our statistical analysis of data distribution and experimental results. They are evaluated on the validation dataset to select a group of parameters with relatively optimal effect.

\subsection{Vehicle Re-ID in single camera}

The single-camera vehicle Re-ID task aims at matching vehicles from the same view in consecutive frames. We utilize SiamRPN++ \cite{article10} as the single tracker model and place such tracker for each target to realize Re-ID in a single camera. Despite the great success of SiamRPN++, we observe that it still fails when the target suffers from a large distortion rate in different positions of the camera, which enlarges differences between target features in different frames. Besides, occlusion between targets leads to more complexity of target location information. Failure cases are shown in Fig. \ref{fig:Figure02}(b). To circumvent the challenges above, we propose the novel post-process method for data association as follows.

\subsubsection{Quality Evaluation Mechanism} 
Unlike the tracking task, the vehicle Re-ID task is less constrained by the bounding box's size and is more sensitive to the bounding box center's drift level. As a consequence, post-processing techniques for tracking outcomes must be tailored to the Re-ID task's needs. We offer a unique quality rating system for monitoring outcomes that is inspired by the attention mechanism.

\textbf{Center Drift} It is essential to update the tracking template to adapt to target movement. $IoU$ and confidence of the output bounding box in consecutive frames are usually taken as indicators to update the template dynamically. Comparatively, the Re-ID process pays more attention to the center drift of the target. In Fig. \ref{fig:Figure02}(b), severe center drift results in many matching errors. However, IoU is not an appropriate metric for the task here. In Fig. \ref{fig:Figure02}(c), the center of the target vehicle is stable while IoU decreases continuously because of the changing size of the bounding box. Updating templates in such circumstances may consume more resources and have a higher risk of wrong predictions. To alleviate this problem, we define a center drift metric, $IoU_R$, to measure the drift level of the target center.

\begin{equation}
IoU_R=\frac{S}{2R^2-S},
\label{Eq(1)}
\end{equation}
where $R$ is the side length of the orange square in the output bounding box center as shown in Fig. \ref{fig:Figure02}(c), we set it as a constant in experiments. $S$ is the intersection area between tracking results of the same target in consecutive frames.

\textbf{Re-ID Confidence.} The confidence score output from the tracking process is used for ranking the bounding boxes. However, it is closely related to the center position and size of the bounding box, which is improper for Re-ID processes. Therefore, we suggest Re-ID confidence ($C_R$) to verify the accuracy of Re-ID results as Eq. (\ref{Eq(2)}).
\begin{equation}
C_R=C_T \times IoU_R,
\label{Eq(2)}
\end{equation}
where $C_T$ is the tracking confidence. The drift level can down weight the scores of bounding boxes far from the previous center of an object. 

We define the conditions for updating the tracking template based on $IoU_R$ and $C_R$ as follows:

\begin{equation}
IoU_{RM}<T_1, C_{RM}<T_2,
\label{Eq(3)}
\end{equation}
where $M$ represents the average result of consecutive $M$ frames, $T_1$, $T_2$ are corresponding thresholds for $IoU_R$ and $C_R$. The template adaptive updating process adapted to the process of the Re-ID task is realized. 

\textbf{Occlusion Coefficient}
To compensate for the box drift induced by occlusion, we incorporate the occlusion coefficient as Eq. (\ref{Eq(4)}):
\begin{equation}
OC=\frac{I_N}{A},
\label{Eq(4)}
\end{equation}

where $I_N$ stands for the intersection area between two tracking results of objects in the same frame, $A$ is the area of the object. An object is defined as severely occluded when $OC$ is greater than the threshold $T_O$. When both objects have high overlapping rates, the object with lower $C_R$ is counted as occluded. We take consecutive frame results as the criterion for dealing with the occluded target as the position relation changes over time. The tracker and ID of the obscured object will be maintained until $N$ consecutive frames of occlusion; otherwise, they would be removed forever.

The $IoU_R$, $C_R$ and $OC$ constitute the quality evaluation mechanism, which processes the tracking results and optimizes the Re-ID performance in a single camera.

\subsubsection{Framework of vehicle Re-ID in single camera}
The overall framework of vehicle Re-ID in a single camera is depicted as Fig. \ref{fig:Figure04}. Each object is allocated a unique tracker, the template of which is populated with the outcome of the object detection. These trackers are responsible for the next frame and provide the quality assessment module with Re-ID findings. A non-qualified result triggers the need to amend the template and initiate another tracking procedure. Depending on the number of consecutive frames, the tracker and ID of obscured objects will be erased or temporarily retained.

	\begin{figure}[thpb]
   \centering
   \parbox{3.3in}
   {\begin{minipage}{1\linewidth}
			\centering
			\centerline{\includegraphics[width=8cm]{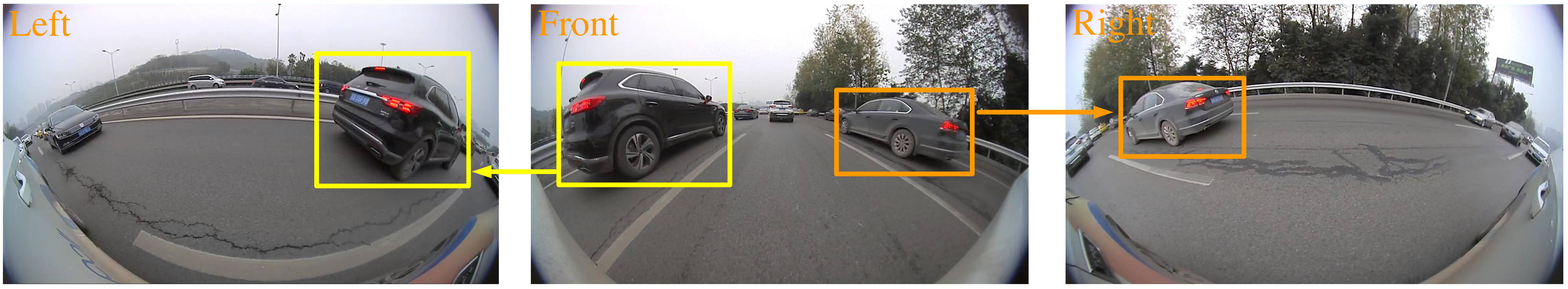}}
			\centerline{(a)}
		\end{minipage}
		
		\begin{minipage}{1\linewidth}
			\centering
			\centerline{\includegraphics[width=8cm]{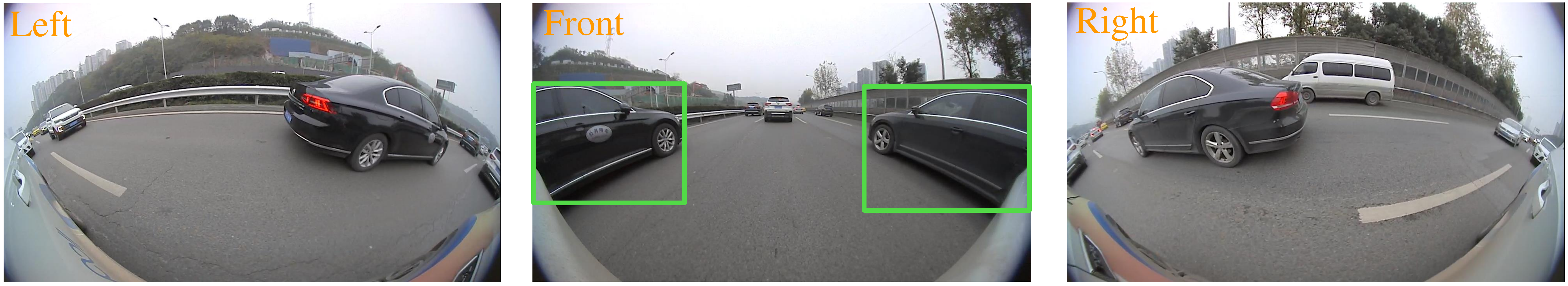}}
			\centerline{(b)}
		\end{minipage}
}
   
   \caption{Samples captured by different cameras. (a) The appearances of the same vehicle captured by different cameras vary greatly, and the same color represents the same object. (b) Objects have a similar appearance may appear in the same camera view, as shown by these two black vehicles in green boxes.}
        \label{fig:Figure05}
\end{figure}

\begin{figure}[!h]
   \centering
   \parbox{3.3in}{\includegraphics[width=7.5cm]{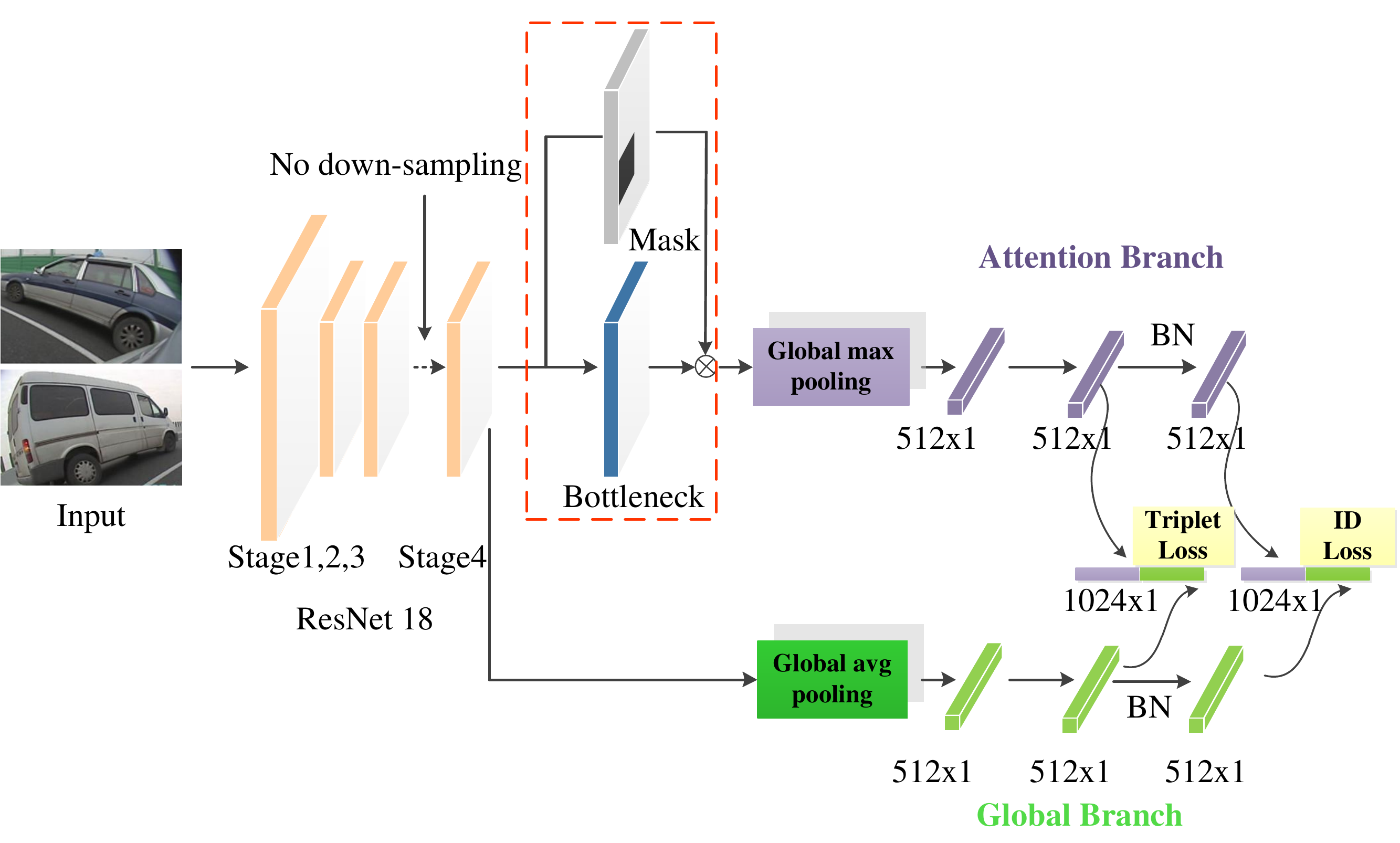}
}
   
   \caption{Illustration of the multi-camera Re-ID network. This network is a two-branch parallel structure. The top branch is employed to make the network pay more attention to object regions, and another is to extract global features.}
    \label{fig:Figure06}
\end{figure}

\subsection{Vehicle Re-ID in multi-camera}
Vehicle Re-ID in multi-camera aims at building correlation between the same vehicle in different cameras. Most of the current methods employ a deep network to achieve Re-ID. However, in a surround-view camera system, cameras are mounted at different positions around the vehicle, resulting in the same object appears variously in different cameras, as shown in Fig. \ref{fig:Figure05}(a). Understanding the fact that adopting general deep learning networks is incapable of handling this problem. Therefore, we introduce an attention module in this paper that forces the network to pay more attention to target areas. Furthermore, different targets that appear in the same camera may have similar appearances, as shown in Fig. \ref{fig:Figure05}(b). It is challenging to distinguish these two black vehicles in front camera only using image-level features. Consequently, we introduce a novel spatial constraint strategy to handle this thorny problem.

\begin{figure}[htbp]
   \centering
   \includegraphics[width=1\linewidth]{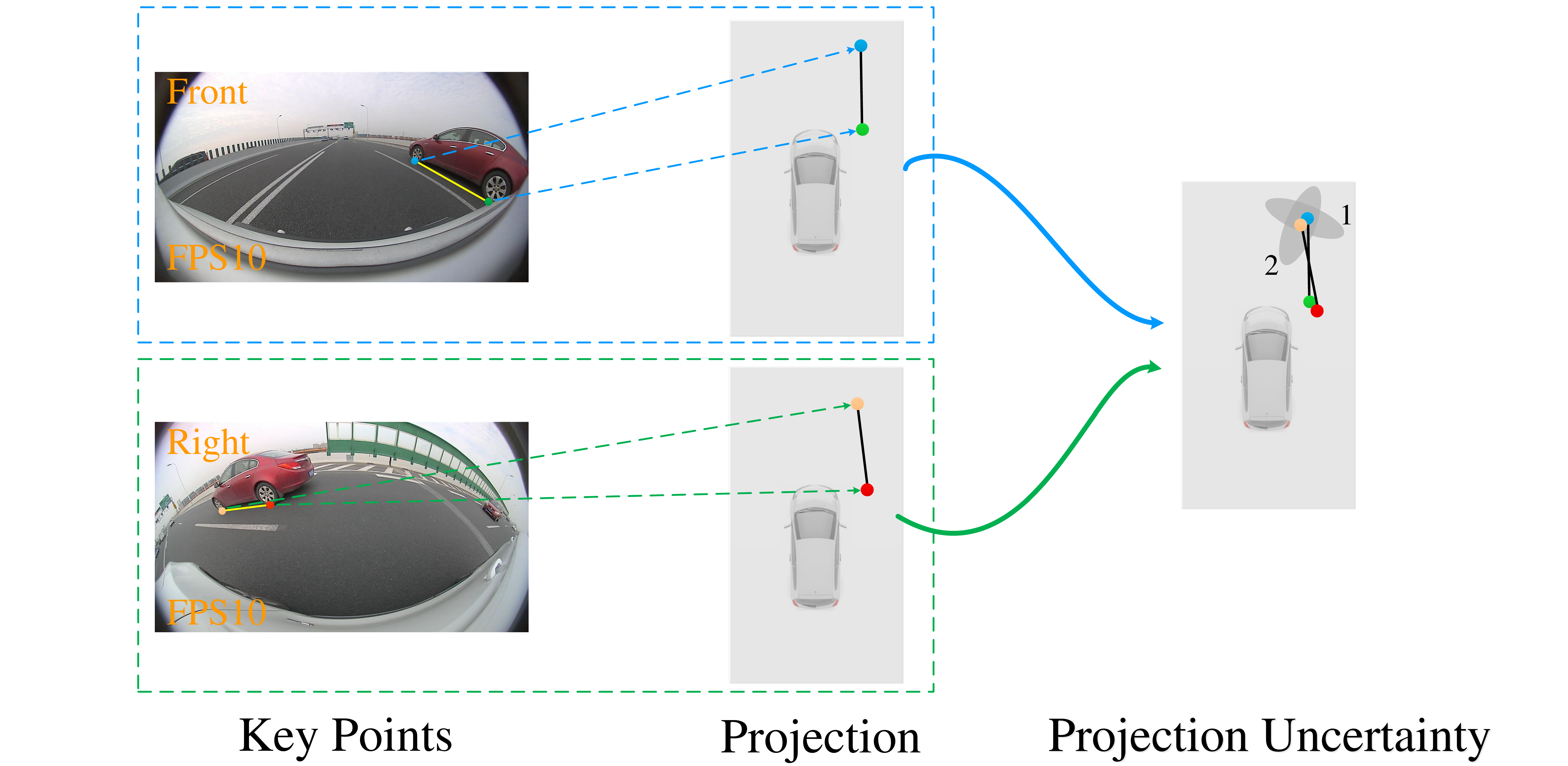}

   \caption{Projection uncertainty of key points. Ellipse 1 and ellipse 2 are uncertainty ranges of front and left (right) cameras, respectively.}
    \label{fig:Figure07}
\end{figure}

\begin{figure*}[!ht]
   \centering
   \includegraphics[width=1\linewidth]{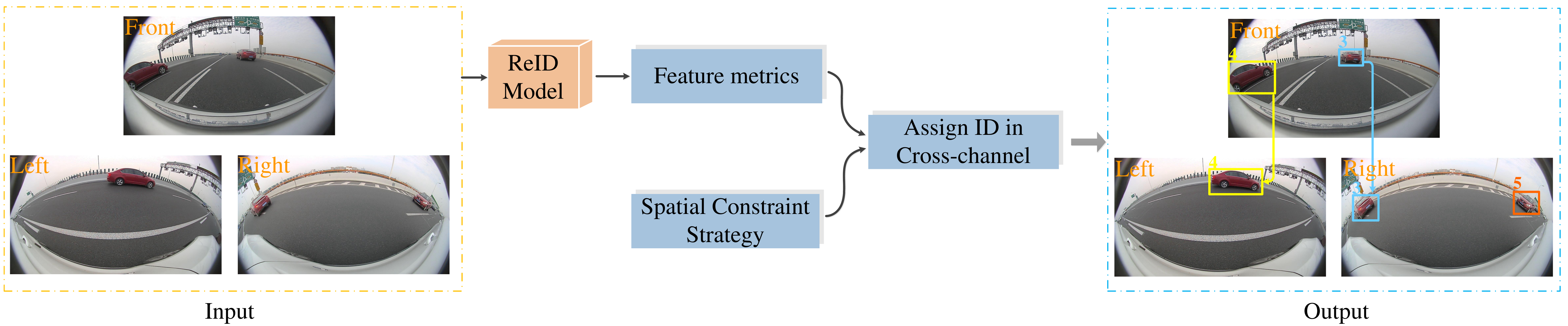}

   \caption{The overall framework of the vehicle Re-ID in multi-camera.For the new target, the Re-ID model is used first to extract the features, followed by the distance metrics is carried out for this feature and features in the gallery. Besides, the spatial constraint strategy is adopted to improve the correlation effect.}
    \label{fig:Figure08}
\end{figure*}

\subsubsection{Attention Module}
Existing vehicle Re-ID methods mainly serve for surveillance scenarios. To meet the requirement of vehicle Re-ID for the surround-view camera system, we apply a modified BDBnet \cite{article11} as our multi-camera Re-ID model. BDBnet consists of a global branch and a local branch. Especially, a fixed mask is added to a local branch to help the network learn semantic features, and it is shown to be effective in pedestrian Re-ID. Different from pedestrian Re-ID, vehicle Re-ID for surround-view camera system suffers from deformation in a multi-camera system. Fixed templates are difficult to improve learning outcomes, so we introduce an attention module that leads the network to learn self-adaptive templates to focus on target regions. As shown in Fig. \ref{fig:Figure06}, the structure in the red box is the attention module. For each new target, the network is applied to extract features and measure the Euclidean distance between this feature and features stored in the feature gallery. Then the distance is converted to confidence score $s_1$ through Eq. (\ref{Eq(5)}). 

\begin{equation}
s_1=ln(\frac{1}{D_F}+1),
\label{Eq(5)}
\end{equation}

\noindent
where $D_F$ is the Euclidean distance between the feature of the new target and features stored in gallery.

\begin{figure*}[!ht]
   \centering
   \includegraphics[width=1\linewidth]{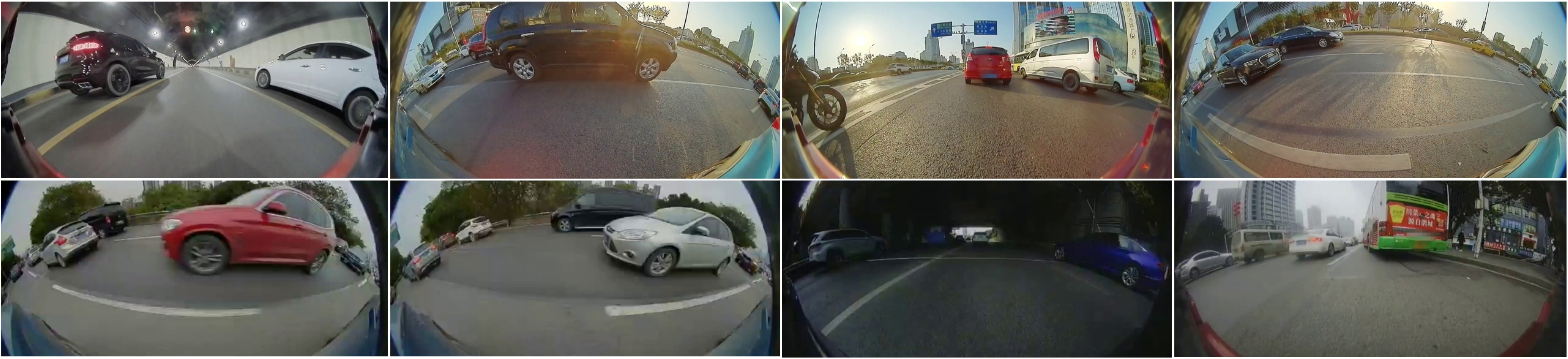}

   \caption{The presentation of the proposed dataset, which contains images of occlusion and illumination.}
    \label{fig:Figure10}
\end{figure*}

\subsubsection{Spatial Constraint Strategy}
We introduce a novel spatial restriction technique for dealing with scenarios in which a single camera contains multiple similar targets. As previously stated, the wheel grounding key points of the same target in multiple cameras correspond to the same real-world coordinate location.
Here we define the wheel grounding key points as the points where the wheel touches the ground. These points are special since their coordinates in the local coordinate system (ego-vehicle coordinate system) are $P(x, y, 0)$. After obtaining the pixel coordinates of the contact points, we can get their physical coordinates in the real world coordinates through the Fisheye IPM algorithm \cite{mallot1991inverse}.
However, the projected position varies due to external factors, such as a camera-mounted angle and calibration. We define the offset caused by these factors as projection uncertainty as shown in Fig. \ref{fig:Figure07}. Two key points of the same category which are projected into the overlapping area are determined to belong to the same vehicle. Furthermore, the error is decreased if the target gets closer to the camera, so we suggest a different standard for wheels in a different position. As presented in Eq. (\ref{Eq(6)}), we first calculate coordinate distances between key points and then convert them to score $s_2$:
\begin{equation}
s_2=ln(\frac{1}{D_K}+1),
\label{Eq(6)}
\end{equation}
where $D_K$ is the distance between the projection coordinate of the key points, and $D_K=D_f+D_r$, $D_f$ and $D_r$ are the projection coordinate distance of these two front key points and two rear key points, respectively.

\subsubsection{Framework of vehicle Re-ID in multi-camera}
The integral process of multi-camera vehicle Re-ID is shown in Fig.\ref{fig:Figure08}. The first branch is used to obtain the confidence score $s_1$ of feature similarity metrics, and the second branch is used to obtain the confidence score $s_2$ of physical coordinate distances of key points. Finally, ID of the target with the highest score $s$ is assigned to the new target.
\begin{equation}
s=\frac{\alpha s_1+\beta s_2}{\alpha+\beta},
\label{Eq(7)}
\end{equation}
where $\alpha$ and $\beta$ are set to 1 in following experiments.

The parameters in Eq. (\ref{Eq(7)}) are based on experimental results, which have been verified on our fisheye dataset. If more accurate wheel position coordinates can be obtained, the weight of physical distance fraction can be appropriately increased, while in the case of large wheel coordinate deviation areas or vehicles with close distances in congestion scenes, the weight of feature similarity can be increased to achieve better results.

The multi-camera ID association in this paper is implemented in a certain order. The association between the left camera and the front camera is carried out first, followed by the association between the front camera and the right camera. In detail, the process is as follows:

1)	For the case that a new target appears on the left side of the left camera or the right side of the right camera, the association only needs to be carried out in their camera. If successful, we assigned the original ID for the new target. If not, we assign a new ID for this target immediately.

2)	When a new target appears on other areas of the left or right camera, it just needs to be associated with the front camera. 

3)	When a new target appears on the right side of the front camera, only the right camera needs to be associated. Similarly, it just needs to be associated with the left camera for the new target appears on the left side of the front camera.

\begin{figure}[ht]   
	\centering
	\includegraphics[width=\linewidth,scale=1.00]{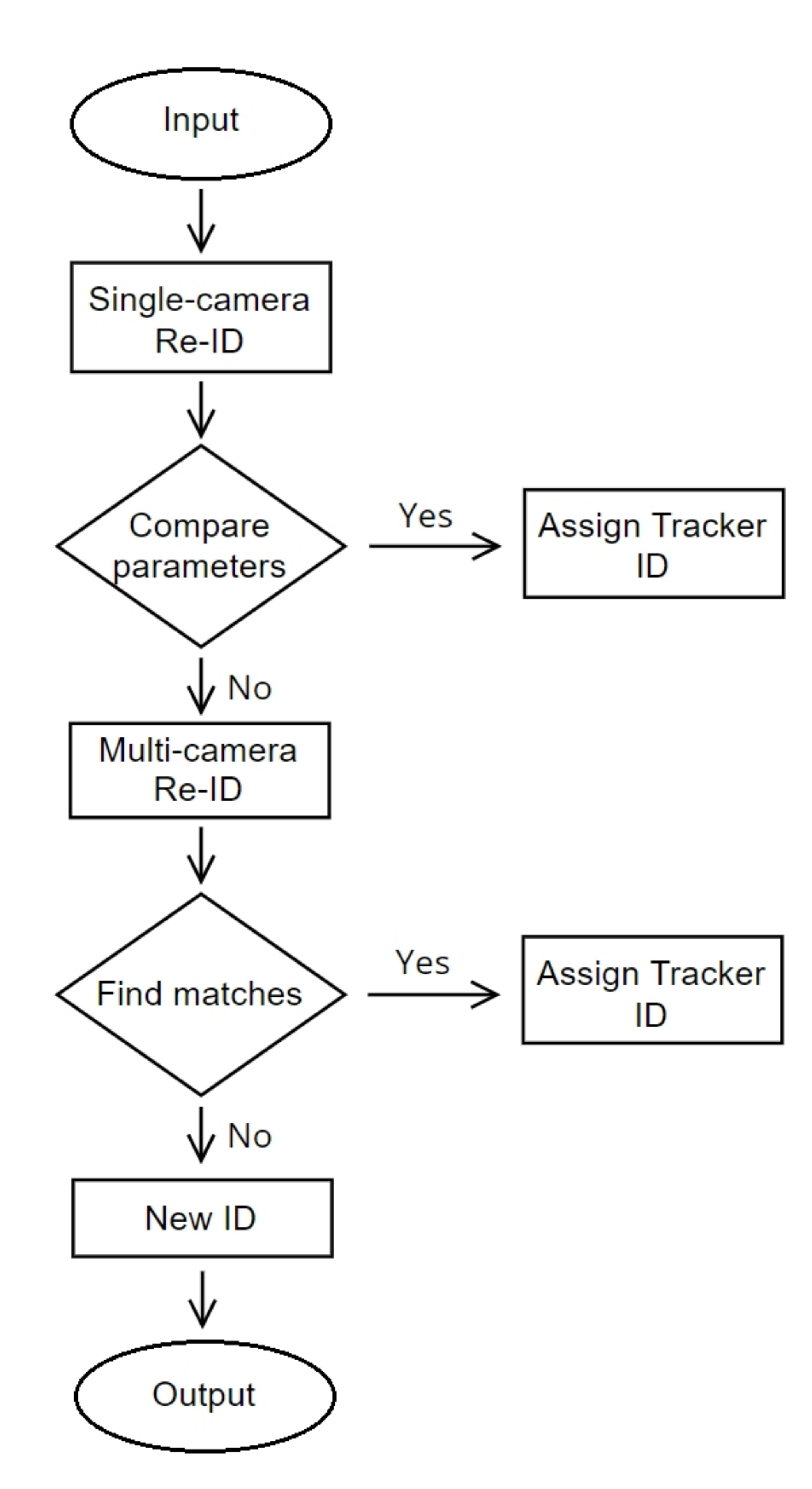}
	\caption{The flowchart of the overall framework. Firstly, each target is attempted to assigned an ID in a single camera. Then the target without ID will be assigned an ID in the multi-camera Re-ID stage when it can be matched. If a target appears for the first time in all camera, it will be assigned an new ID. Finally, every target has an ID and the same target in different cameras has the same ID.}
	\label{fig:Figure09}
\end{figure}

\subsection{Overall framework}
The surround-view multi-camera Re-ID system processes data of each camera serially. Each new object in a single channel will be assigned a single target tracker and matched with data of other channels according to the Re-ID strategy in multi-camera. If not matched successfully, a new ID will be created for it. Else, the ID of the matching object will be inherited by the newcomer. According to the Re-ID strategy in a single camera, all targets in a single camera will be matched in time sequence.

To facility the representation in engineering applications, it is necessary to use a vector to describe a target during single and multi-camera Re-ID stages. In the single-camera stage, the target-vehicle vector comprises the object id and tracker information such as the bounding boxes $(X, Y, W, \\H)$ information and occlusion coefficient, which are independent in each camera. Then the target vector with id generated by the Re-ID method in the single-camera stage is sent to the multi-camera Re-ID stage. Since the same target may be observed in different channels, the goal of the multi-camera Re-ID stage is to fuse the same targets between channels and update the same identification numbers in the vectors. In this stage,  new targets information, such as the wheel coordinates, can be generated and appended to the vectors. Finally, we could use vectors to represent all targets. Each vector contains a target's complete information, including a unique identity number. The overall flow of the model is shown in the flowchart in Fig.\ref{fig:Figure09}

\section{Experiments}
\subsection{Dataset and evaluation metric}

We generate the fisheye dataset from structured roads. It consists of a total of 22190 annotated images sampled from 36 sequences. Our dataset contains a large number of images with occluded and illuminated scenes, as shown in Fig.\ref{fig:Figure10}. The resolution of the image is 1280$\times$720. We use the 80$\%$ and 20$\%$ images as training and testing sets.
We utilize three cameras (i.e., front, left, and right) to capture these images at 30 frames per second (fps). The dataset is divided into amounts of image sequences. We select a group of image sequences that have about 80$\%$ of the total number of images and use these images as training sets. The rest of the image sequence serves as the verification set. The comparison with other approaches is shown in Table \ref{table04}. They are deployed on the Qualcomm 820A platform
with an Adreno 530 GPU and a Hexagon 680 DSP.
The results prove our proposed approach is computationally efficient.

\begin{table}[t]
\setlength{\tabcolsep}{21pt}
\caption{The comparison with other approaches on speed. }\label{table04}
\begin{center}
\begin{tabular}{ll}
\hline\noalign{\smallskip} 
 Methods  & Speed(FPS)\\
\noalign{\smallskip}\hline\noalign{\smallskip}

VANET[\cite{chu2019vehicle}]   & \quad\ \ \ 19    \\
SPAN[\cite{chen2020orientation}]    & \quad\ \ \ 13    \\
PVEN[\cite{meng2020parsing}]    & \quad\ \ \ 12    \\
Ours    & \quad\ \ \ 30    \\

\noalign{\smallskip}\hline
\end{tabular}
\end{center}

\end{table}

\begin{table}[h]
\setlength{\tabcolsep}{17pt}
\caption{The impact of the different methods updating tracking template. $IC_{front}$, $IC_{left}$ and $IC_{right}$ correspond to the $IC$ of front, left and right cameras. }\label{table01}
\begin{center}
\begin{tabular}{llll}
\hline\noalign{\smallskip} 
 Methods & \begin{tabular}[c]{@{}c@{}}$IC_{front}$\\ \end{tabular} & \begin{tabular}[c]{@{}c@{}}$IC_{left}$\\
 \end{tabular} & \begin{tabular}[c]{@{}c@{}}$IC_{right}$\\ 
 \end{tabular} \\
\noalign{\smallskip}\hline\noalign{\smallskip}

Default         & 0.82      & 0.81     & 0.87\\
+$IoU_T+C_T$    & 0.88      & 0.91     & 0.90\\
+$IoU_R+C_R$    & 0.94      & 0.96     & 0.97\\

\noalign{\smallskip}\hline
\end{tabular}
\end{center}

\end{table}

We evaluate the results with the identity consistency ($IC$). It is formulated as 

\begin{equation}
IC=1-\frac{\sum_tIDSW_t}{\sum_tID_t},
\label{Eq(8)}
\end{equation}
\noindent
where $t$ is the frame index, identity switch (IDSW) is counted if a ground truth target $i$ is matched to tracking output $j$ and the last known assignment is $k$, $k\neq j$. $ID_t$ is the sum of ground truth targets with ID in frame $t$.

\subsection{Implementation Details}
We trained all networks with stochastic gradient descent (SGD) on a GTX 1080Ti.

For the single-camera Re-ID model SiamRPN++, 50 epochs with batch size 24 were trained with a learning rate of 0.0005, weight decay of 0.0001 and momentum of 0.9. For the multi-camera Re-ID model, 150 epochs with batch size 256 were trained with an initial learning rate of 0.00035 (fix for first ten epochs), weight decay of 0.9 and momentum of 0.0005. The hyperparameters values of the network, such as learning rate and batch size, are based on a group of parameters obtained by empirical tuning. The values of $\alpha$ and $\beta$ are obtained experimentally, is shown in Table \ref{table07}

\begin{table}[h]

\setlength{\tabcolsep}{11pt}
\caption{The effect of setting the values of $\alpha$ and $\beta$ on $IC_{front}$, $IC_{left}$ and $IC_{right}$ in Eq. (\ref{Eq(7)}).}\label{table07}
\begin{center}
\begin{tabular}{cccc}
\hline\noalign{\smallskip} 
   & $\alpha$=1, $\beta$=1 & $\alpha$=2, $\beta$=1 & $\alpha$=1, $\beta$=2 \\
\noalign{\smallskip}\hline\noalign{\smallskip}

\begin{tabular}[c]{@{}c@{}}$IC_{front}$\\ \end{tabular} & 0.94      & 0.95     & 0.90\\
\begin{tabular}[c]{@{}c@{}}$IC_{left}$\\ \end{tabular} & 0.96      & 0.94     & 0.96\\
\begin{tabular}[c]{@{}c@{}}$IC_{right}$\\ \end{tabular} & 0.97      & 0.97     & 0.93\\

\noalign{\smallskip}\hline
\end{tabular}
\end{center}

\end{table}

\subsection{Evaluation of the proposed method}
\textbf{Quality Evaluation Mechanism.} The proposed quality evaluation mechanism is a key component to optimize Re-ID performance in a single camera. Therefore, we conduct some ablation studies on our dataset to find out the contribution of our method to performance.

\begin{table}[h]
\setlength{\tabcolsep}{17pt}
\caption{The comparison of our best result and other similar approaches on IC. }\label{table05}
\begin{center}
\begin{tabular}{llll}
\hline\noalign{\smallskip} 
 Approaches  & \begin{tabular}[c]{@{}c@{}}$IC_{front}$\\ \end{tabular} & \begin{tabular}[c]{@{}c@{}}$IC_{left}$\\
 \end{tabular} & \begin{tabular}[c]{@{}c@{}}$IC_{right}$\\ 
 \end{tabular} \\
\noalign{\smallskip}\hline\noalign{\smallskip}

VANET       & 0.90      & 0.90     & 0.91\\
SPAN        & 0.91      & 0.93     & 0.92\\
PVEN        & 0.93      & 0.94     & 0.95\\
Our         & 0.94      & 0.96     & 0.97\\

\noalign{\smallskip}\hline
\end{tabular}
\end{center}

\end{table}

We first compare the template updating metrics. As shown in Table \ref{table01}, Default is our baseline model without updating templates. Updating the tracking template with $IoU_T$ and $C_T$ as metrics brings significant improvement in identity consistency. Furthermore, when utilizing the $IoU_R$ and $C_R$, the identity consistency is greatly improved once again. That means the revised metrics help the model pay more attention to the demands of the Re-ID task. To further evaluate the proposed method, we compared our model with other approaches as shown in Table \ref{table05}. Meanwhile, we also present the number of parameters in Table \ref{table06}, which shows our proposed method is cost-effective.

\begin{table}[h]
\setlength{\tabcolsep}{21pt}
\caption{The comparison with other approaches on amount of parameters. }\label{table06}
\begin{center}
\begin{tabular}{ll}
\hline\noalign{\smallskip} 
 Methods  & Parameters(M)\\
\noalign{\smallskip}\hline\noalign{\smallskip}

Satt[\cite{liu2019urban}]    & \quad\ \ \ 31.09    \\
GSTE[\cite{ma2019vehicle}]    & \quad\ \ \ 32.73    \\
VANET   & \quad\ \ \ 10.9    \\
HPGN[\cite{shen2021exploring}]    & \quad\ \ \ 27.71    \\
Ours    & \quad\ \ \ 8.19    \\

\noalign{\smallskip}\hline
\end{tabular}
\end{center}

\end{table}
The number of frames ($N$) to delete the occluded object are summarized in Table \ref{table02}. Zero frame means that we do not handle the occluded cars. When $N=2$, the identity switch is decreased slightly. It is expected because deleting the temporarily occluded cars frequently contributes to the $ID$ switch as its rapid reappearance. As $N$ grows, the $ID$ is gradually steady. However, maintaining too many frames means don’t handle occlusion, and the consistency descends once again. Experimental results show that $N = 4$ is optimal in this paper.\newline

\noindent
\textbf{Discussion of robustness.} The fisheye dataset verifies our proposed system's robustness under different illumination and occlusion. The dataset was captured in different daytime periods, including diverse illumination and various occlusion conditions caused by different traffic conditions such as traffic jam scenarios. Table \ref{table02} shows the robustness of our method in an occlusion situation. Besides, the frame rate can reach about 30 fps in practical tests, which can meet real-time demand. However, it should also be noted that our dataset does not include rainy and night scenes.

\begin{table}[h]
\setlength{\tabcolsep}{21pt}
\caption{The impact of the different methods updating tracking template. }\label{table02}
\begin{center}
\begin{tabular}{llll}
\hline\noalign{\smallskip} 
 N & \begin{tabular}[c]{@{}c@{}}$IC_{front}$\\ \end{tabular} & \begin{tabular}[c]{@{}c@{}}$IC_{left}$\\
 \end{tabular} & \begin{tabular}[c]{@{}c@{}}$IC_{right}$\\ 
 \end{tabular} \\
\noalign{\smallskip}\hline\noalign{\smallskip}

0   & 0.83      & 0.81     & 0.80\\
2   & 0.85      & 0.83     & 0.84\\
3   & 0.92      & 0.90     & 0.93\\
4   & 0.94      & 0.96     & 0.97\\
5   & 0.93      & 0.91     & 0.92\\

\noalign{\smallskip}\hline
\end{tabular}
\end{center}

\end{table}

\noindent
\textbf{Vehicle Re-ID strategy in multi-camera.} We examine the influence of various matching strategies in multi-camera in Table \ref{table03}. All the experiments are based on the same implementations in a single view. The first row shows the method of matching with feature metrics as \cite{article11}. After introducing the attention module, the Re-ID accuracy has achieved a promising improvement. Based on that, adding the spatial constraints strategy improves, as shown in the last row.

\begin{table}[h]
\setlength{\tabcolsep}{6pt}
\caption{Ablation study in multi-camera Re-ID evaluated on proposed fisheye dataset.}\label{table03}
\begin{center}
\begin{tabular}{llll}
\hline\noalign{\smallskip} 
 ID Matching strategy in multi-camera  & \begin{tabular}[c]{@{}c@{}}$IC_{front}$\\ \end{tabular} & \begin{tabular}[c]{@{}c@{}}$IC_{left}$\\
 \end{tabular} & \begin{tabular}[c]{@{}c@{}}$IC_{right}$\\ 
 \end{tabular} \\
\noalign{\smallskip}\hline\noalign{\smallskip}

Baseline                      & 0.84      & 0.85     & 0.86\\
+Attention module             & 0.87      & 0.92     & 0.89\\
\begin{tabular}[c]{@{}l@{}}+Spatial constraint strategy  \\and Attention module\end{tabular}
& 0.94      & 0.96     & 0.97 \\

\noalign{\smallskip}\hline
\end{tabular}
\end{center}

\end{table}

\noindent
\subsection{Evaluation of positioning error using in spatial constraint strategy }
\begin{figure}[htbp]
\centering
\subfigure[]
{
\begin{minipage}{3.9cm}
\centering
\includegraphics[width=3.9cm]{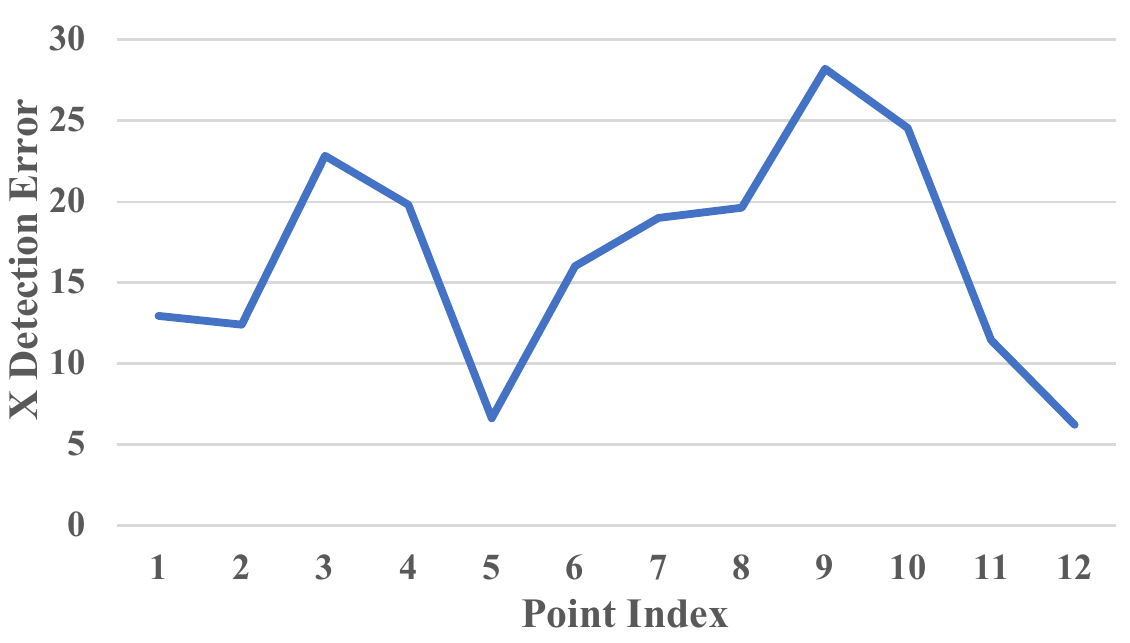}
\newline
\end{minipage}
}
\subfigure[]
{
\begin{minipage}{3.9cm}
\centering
\includegraphics[width=3.9cm]{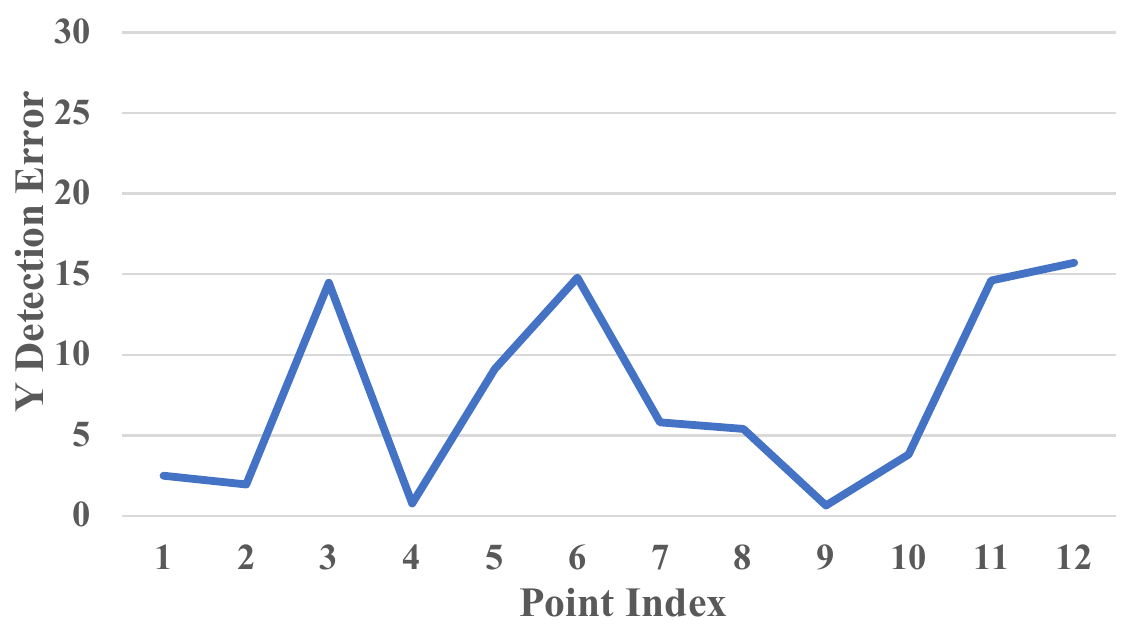}
\newline
\end{minipage}
}

\caption{Evaluation of the positional accuracy on the ground with 5$\%$ slope gradient. Best view in color and zoom in.}
\label{fig_position_err}
\end{figure}

We conduct an experiment to evaluate the object position accuracy, which affects the effectiveness of the spatial constraint strategy. Since there is no ideal horizontal ground plane in practical application, we designed an experiment to evaluate the positioning accuracy applied in constraint strategy on the ground with a 5$\%$ slope gradient. We randomly select 12 objects distributing from -2.5 to 2.5 meters along $x$-direction and calculate their position errors. As shown in Fig.\ref{fig_position_err} (a), the result demonstrates that all the errors are less than 30 cm. Meanwhile, 12 points are randomly selected in the range of 1.5 to 3.5 meters along $y$-direction to analyze the position errors. As shown in Fig.\ref{fig_position_err} (b), the position errors of objects are less than 20 cm. It can be concluded that our system has the capacity to work on the ground with a slight slope gradient, which proves the robustness of the proposed system in practical scenarios.

\section{Conclusions}
We present a comprehensive vehicle Re-ID approach for the surround-view multi-camera situation in this study. The introduced quality evaluation mechanism for the output bounding box can help eliminate distortion and occlusion-related issues. Moreover, we deploy an attention module to direct the network's attention to certain locations. Additionally, a unique spatial restriction method is applied in this situation to regularise the Re-ID findings greatly. Extensive component analysis and comparisons on the fisheye dataset demonstrate that our vehicle Re-ID solution produces promising results. Our model achieved 30 FPS on the Qualcomm 820A platform and its number of parameters is only 8.19 million. Furthermore, the annotated fisheye dataset will be made publicly available to aid in advancing research in this area. We will continue to optimise the performance of vehicle Re-ID in the surround-view multi-camera situation in future investigations.
\bibliographystyle{spbasic}      
\bibliography{references} 
%
%

\end{document}